\documentclass[10pt, letterpaper, conference]{ieeeconf}

\IEEEoverridecommandlockouts                              

\usepackage[]{graphicx}
\usepackage[cmex10]{amsmath}
\usepackage{amsfonts, amssymb}
\usepackage{mdwmath}
\usepackage{mdwtab}
\usepackage{multirow}
\usepackage{subcaption}
\newsavebox{\bigimage}
\usepackage{algorithm}	
\usepackage{algpseudocode}
\usepackage{balance}
\usepackage{comment}
\usepackage{gensymb}
\usepackage{hyperref}

\let\oldhat\hat

\renewcommand{\hat}[1]{\oldhat{\mathbf{#1}}}

\newdimen\figrasterwd
\figrasterwd\textwidth

\title{\LARGE \bf
Accurate 3D Localization for MAV Swarms by UWB and IMU Fusion
}

\author{Jiaxin Li$^{1}$, Yingcai Bi$^{1}$, Kun Li$^{2}$, Kangli Wang$^{2}$, Feng Lin$^{3}$, Ben M. Chen$^{2}$
\thanks{$^{1}$NUS Graduate School for Integrative Science \& Engineering, National University of Singapore (NUS). 
        {\tt\small jli@u.nus.edu}}%
\thanks{$^{2}$Department of Electrical \& Computer Engineering, NUS}%
\thanks{$^{3}$Temasek Lab, NUS.}%
}

\begin{document}

\maketitle

\begin{abstract}
Driven by applications like Micro Aerial Vehicles (MAVs), driver-less cars, etc, localization solution has become an active research topic in the past decade. In recent years, Ultra Wideband (UWB) emerged as a promising technology because of its impressive performance in both indoor and outdoor positioning. But algorithms relying only on UWB sensor usually result in high latency and low bandwidth, which is undesirable in some situations such as controlling a MAV. To alleviate this problem, an Extended Kalman Filter (EKF) based algorithm is proposed to fuse the Inertial Measurement Unit (IMU) and UWB, which achieved 80Hz 3D localization with significantly improved accuracy and almost no delay. To verify the effectiveness and reliability of the proposed approach, a swarm of 6 MAVs is set up to perform a light show in an indoor exhibition hall. Video and source codes are available at \url{https://github.com/lijx10/uwb-localization}
\end{abstract}

\section{Introduction}
Micro Aerial Vehicle (MAV) swarm has been attracting increasing interest in recent years because of its potential improvements on a variety of applications, such as surveillance \cite{nigam2012control}, communication relay \cite{hayat2016survey}, 3D reconstruction \cite{haris2017super} and so on. To control a formation of MAV swarm, accurate localization is at the forefront. Usually the localization is realized by global navigation satellite system (e.g. GPS) \cite{dong2016time}, or by high-precision optical tracking system (e.g. VICON) \cite{kushleyev2013towards}. The accuracy of standard GPS is meter-level and it can be boosted to centimeter-level, when running as Real Time Kinetic (RTK) mode with an additional fixed base station. GPS can only work in open space and the accuracy is susceptible to poor satellite signal quality due to multi-path effect. On the other hand, VICON system provides centimeter-level accuracy in indoor environment but it requires an expensive and complex setup with multiple cameras.

There are more and more attempts on alternative localization solutions for MAV swarm without GPS and expensive VICON system. Vision based solution is of particular interest because it is low-cost, light-weight and flexible. In \cite{nageli2014environment} a formation estimator is developed by incorporating a consensus-based distributed Kalman filter. An onboard camera is used to track markers and measure relative pose between MAVs. Common features of the environment are shared by different MAVs to estimates the relative pose between them in \cite{montijano2016vision}. The images are first rectified with compensation from IMU rotation and relative pose is calculated based on homography decomposition. In \cite{neunert2016open} a visual-inertial localization system is proposed based on IMU measurements and AprilTag markers. The AprilTags are tracked robustly and Extended Kalman Filter (EKF) is applied to estimate both camera and marker pose simultaneously. IMU is used to improve system robustness in the case of marker detection lost in a short period. Despite of the advantages of visual localization, heavy computational power and proper lighting condition are necessary, which limits its applications.

Radio based localization systems, such as Radio Frequency Identification (RFID) \cite{zhou2009rfid}, WiFi \cite{polo2014semantic}, Zigbee \cite{watthanawisuth2014design} and Ultra-Wideband (UWB) \cite{alarifi2016ultra}, are emerging technologies in indoor positioning. RFID is widely used in logistic management for labeling and tracking assets. By reading a tag with predefined location, the tracking position can be determined with proximity and received signal strength (RSS) value \cite{ruiz2012accurate}. However, RFID can only work with a small proximity range and ranging with RSS is only a rough estimate. WiFi and Zigbee are both configured as wireless network system. They can be deployed for positioning by evaluating the distance based on RSS value and known distribution of network nodes. WiFi is easily available with low cost, and Zigbee is designed with low power consumption. WiFi is likely to be interfered by commonly used mobile devices, which results in meter-level accuracy, which is not good enough for controlling a swarm of UAVs. Zigbee improves the positioning accuracy to some extent \cite{syut2013ombining}, but it is still difficult to maintain a collision-free MAV formation in limited indoor space.

Compared with the above solutions, UWB is one of the most promising solutions in terms of accuracy, coverage range and deployment cost. Thanks to the miniature of single chip transceiver \cite{decawaveweb}, it can be applied as positioning system for real-time robot applications \cite{krishnan2007uwb}. The range between two UWB nodes can be determined by the time difference of arrival (TDOA) \cite{xu2006position, fan2017data}.  At the same time, high capacity data transmission can be implemented with small energy consumption. Multi-path interference is reduced due to the high bandwidth and short-pulse waveform. Moreover, these characteristics can help signals to pass through obstacles and improve the ranging accuracy. Therefore, UWB is an excellent high-accuracy positioning system for MAV swarms in GPS-denied environments. 

\section{Sensor Setup} \label{sec_sensor_setup}
\begin{figure}[t!] \centering
    \includegraphics[width=0.4\columnwidth]{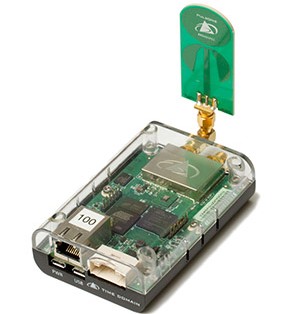}
    \caption{The UWB module}\label{fig_p440}
\end{figure}

\begin{figure}[th!] \centering
\includegraphics[width=0.9\columnwidth]{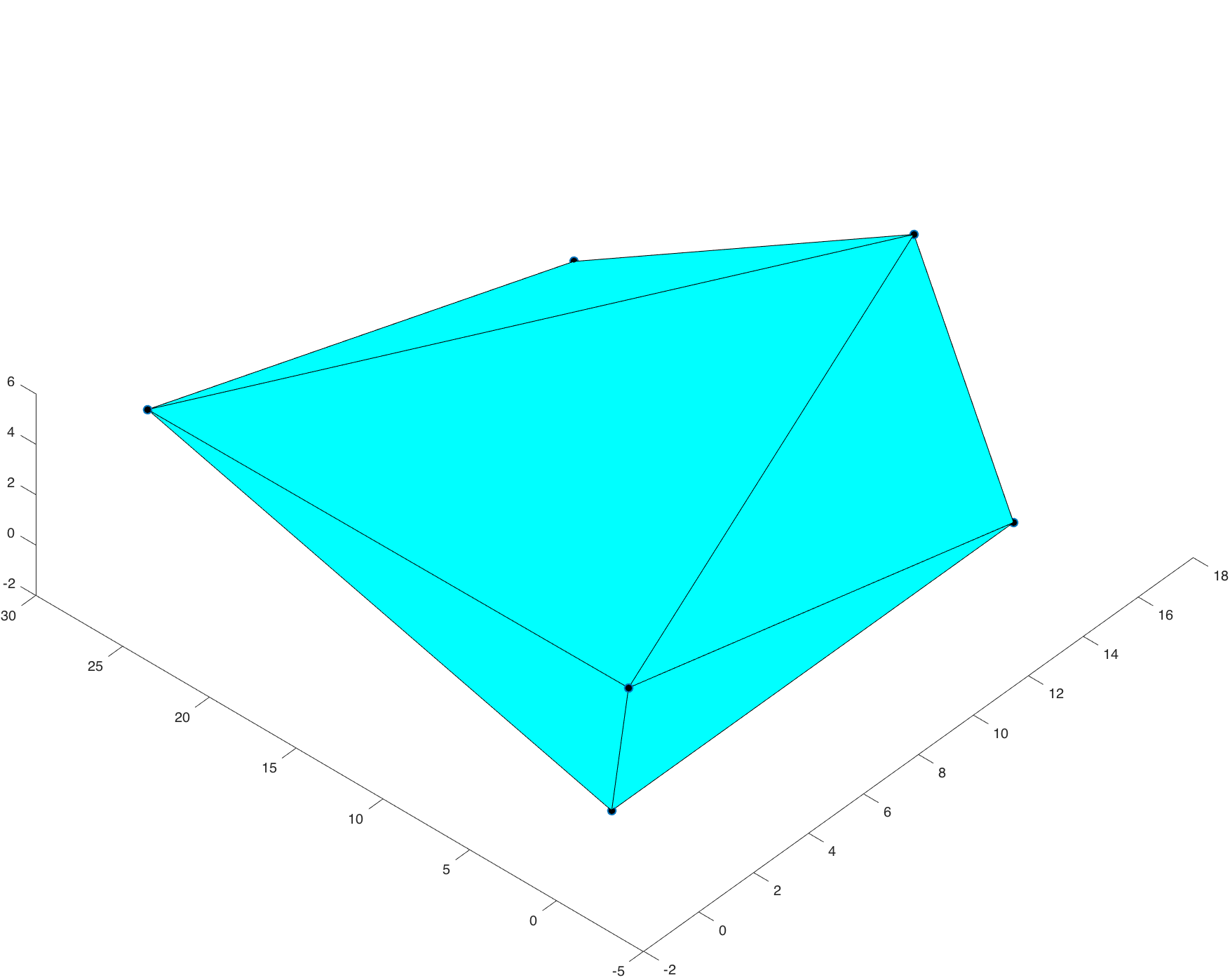}
\caption{UWB anchor placement at Changi Exhibition Centre, Singapore.}\label{fig_anchor}
\end{figure}

In our application of MAV swarm, multiple UWB sensors manufactured by TimeDomain (shown in Fig. \ref{fig_p440}) are set up as a two-way ranging system. 6 UWB sensors are fixed in the exhibition hall as anchors. The anchors enclose roughly a cuboid, and we build a coordinate system of the anchors by precisely measuring the distances and angles between them. Our anchor setup at the Changi Exhibition Centre, Singapore is shown in Fig. \ref{fig_anchor} and (\ref{equ_anchor}), using North-West-Up (NWU) coordinates. The details can be found in Section \ref{sec_experiments}. In Fig. \ref{fig_anchor}, the localization algorithms can achieve utmost performance inside the convex hull, which is a well proved knowledge in the community of sensor placement.

\begin{equation} \label{equ_anchor}
\begin{split}
    & p_{Ax0}=0, p_{Ay0}=0, p_{Az0}=0 \\
    & p_{Ax1}=14.6, p_{Ay1}=0, p_{Az1}=0 \\
    & p_{Ax2}=14.6, p_{Ay2}=25.5, p_{Az2}=0 \\
    & p_{Ax3}=0, p_{Ay3}=-1, p_{Az3}=5.3 \\
    & p_{Ax4}=0, p_{Ay4}=26.6, p_{Az4}=5.3 \\
    & p_{Ax5}=17.4, p_{Ay5}=10.1, p_{Az5}=5.3 
\end{split}
\end{equation}

The two-way ranging system, which is also called Time-of-Arrival (TOA) system, is configured with the techniques of Time Division Multiple Access (TDMA) and Code Division Multiple Access (CDMA). Each MAV is equipped with an UWB sensor, denoted as mobile, that works at a unique channel. In each time slot, each mobile initiates a message to an anchor and the anchor responses immediately. Since the time of all the UWB sensors are synchronized to the precision of picosecond, each mobile can calculate the distance to the anchor it talks to, according to their timing of receiving the other's UWB pulse. The mobiles probe the anchors in a round-robin fashion, i.e., in every 6 time slots, a mobile will probe all 6 anchors and get the respective distance measurements $r_k$ and the measurement uncertainty $\sigma_{rk}$. In our setup, each mobile gets 80 measurements per second.

\section{Vanilla EKF for UWB} \label{sec_vanilla_EKF}
Typical TOA based algorithms include multi-dimensional scaling (MDS), weighted least square, trilateration, etc. These approaches require at least 3 measurements to determine the 3D position, but in a typical UAB setup, each mobile receives one distance measurement at a time, which means that any three measurements are acquired with different time-stamps. On the other hand, filtering based approaches, such as EKF and particle filter, are able to consume one measurement at a time. However, without other information, they are usually under some motion assumptions like constant velocity model, which leads to inaccurate or delayed positioning for MAVs.

In a vanilla EKF, the state vector consists of position and velocity defined in (\ref{equ_vanilla_state}). Under the constant velocity model, the time update process is defined in (\ref{equ_time_update}).
\begin{equation} \label{equ_vanilla_state}
x = [p_x, v_x, p_y, v_y, p_z, v_z]^T \in \mathbb{R}^6
\end{equation}
\begin{equation} \label{equ_time_update}
\begin{split}
\bar{x}_{k} &= A_k x_{k-1} + B_{k-1}u_{k-1} \\
\bar{P}_k &= A_k P_{k-1} A_k^T + Q_{k-1}
\end{split}
\end{equation}
where the control input $u_{k-1}$ and input gain $B_{k-1}$ are zero. System matrix $A_k$ can be derived from the constant velocity assumption, while the process noise covariance $Q_{k-1}$ is derived with the assumption that the acceleration is a Gaussian distribution $\mathcal{N}(0, \sigma_a)$. In this paper, we set $\sigma_a=0.125$ empirically. $T$ is the time interval between each EKF iteration.
\begin{equation} \label{equ_vanilla_A}
A_k = 
\begin{bmatrix}
A'_k & 0 & 0 \\
0 & A'_k & 0 \\
0 & 0 & A'_k
\end{bmatrix}
,
A'_k =
\begin{bmatrix}
1 & T \\
0 & 1
\end{bmatrix}
\end{equation}
\begin{equation} \label{equ_vanilla_Q}
Q_{k-1} = 
\begin{bmatrix}
Q'_{k-1} & 0 & 0 \\
0 & Q'_{k-1} & 0 \\
0 & 0 & Q'_{k-1}
\end{bmatrix}
,
Q'_{k-1} = 
\begin{bmatrix}
\frac{\sigma_a T^4}{3} & \frac{\sigma_a T^3}{2} \\
\frac{\sigma_a T^3}{2} & \frac{T}{2}
\end{bmatrix}
\end{equation}

Denote the position of an anchor as $p_{Ax}, p_{Ay}, p_{Az}$, the predicted distance between mobile and anchor is,
\begin{equation} \label{equ_predicted_r}
\bar{r}_k = \sqrt{(p_x-p_{Ax})^2+(p_y-p_{Ay})^2+(p_z-p_{Az})^2}
\end{equation}

The measurement matrix $H_k$ is defined by first order Taylor expansion of (\ref{equ_predicted_r}),
\begin{equation} \label{equ_H}
H_k =
\begin{bmatrix}
\frac{p_x-p_{Ax}}{\bar{r}_k} & 0 & \frac{p_y-p_{Ay}}{\bar{r}_k} & 0 & \frac{p_z-p_{Az}}{\bar{r}_k} & 0
\end{bmatrix}^T
\end{equation}

The measurement update of EKF is defined below, where the measurement is denoted as $r_k$ and measurement noise covariance is $R_k=\sigma_r^2$. $\sigma_r$ represents the measurement uncertainty between a pair of mobile and anchor which is reported by the UWB sensors.
\begin{equation} \label{equ_measurement_update}
\begin{split}
K_k &= \bar{P}_k H_k^T (H_k \bar{P}_k H_k^T + R_k)^{-1} \\
x_k &= \bar{x_k} + K_k (r_k - \bar{r}_k) \\
P_k &= (I-K_k H_k)\bar{P}_k
\end{split}
\end{equation}
Equations (\ref{equ_vanilla_state}-\ref{equ_measurement_update}) define a vanilla EKF for UWB localization, under the constant velocity assumption. In situations when the sensor is with low acceleration rate, the vanilla EKF performs well, while severe delay can be observed in applications with MAVs, as demonstrated in Section \ref{sec_experiments}.

\section{UWB \& IMU Fusion} \label{sec_fusion_EKF}
In order to solve the delay and low bandwidth problem brought by constant velocity assumption, fusing the acceleration information from IMU is an ideal solution. However, the acceleration measurements from low-cost commercial IMUs are extremely noisy. Moreover, the acceleration bias is unstable, i.e., its bias is affected by many factors including temperature, operating duration, mechanical vibration, etc. Directly integrating the acceleration from IMU may lead to even worse result than that of constant velocity assumption.

\subsection{Augmented State Vector}
To remedy the bias problem, we augment the state vector with acceleration bias in all three axis, i.e. building an augmented state vector $x\in \mathbb{R}^9$. The acceleration measurements $[a_x, a_y, a_z]^T$ serve as the input of (\ref{equ_time_update}). Accordingly, the system matrix $A$, input gain $B_{k-1}$ and $u_{k-1}$ are defined in (\ref{equ_acc_A}-\ref{equ_acc_u}).
\begin{equation} \label{equ_acc_state}
x = [p_x, v_x, a_{bx}, p_y, v_y, a_{by}, p_z, v_z, a_{bz}]^T \in \mathbb{R}^9
\end{equation}
\begin{equation} \label{equ_acc_A}
A_{k} = 
\begin{bmatrix}
A'_k & 0 & 0 \\
0 & A'_k & 0 \\
0 & 0 & A'_k
\end{bmatrix}
,
A'_{k-1} = 
\begin{bmatrix}
1 & T & \frac{-T^2}{2} \\
0 & 1 & -T \\
0 & 0 & 1 
\end{bmatrix}
\end{equation}

\begin{equation} \label{equ_acc_B}
B_{k-1} = 
\begin{bmatrix}
B'_{k-1} & 0 & 0 \\
0 & B'_{k-1} & 0 \\
0 & 0 & B'_{k-1}
\end{bmatrix}
,
B'_{k-1} = 
\begin{bmatrix}
\frac{T^2}{2} & 0 & 0 \\
T & 0 & 0 \\
0 & 0 & 0
\end{bmatrix}
\end{equation}

\begin{equation} \label{equ_acc_u}
u_{k-1} = [a_x, 0, 0, a_y, 0, 0, a_z, 0, 0]^T
\end{equation}
With the assumption that IMU readings are corrupted with Gaussian noise, we use $\tau_a$ to measure the IMU noise. Similarly, $\tau_b$ is used to measure the uncertainly of the estimated acceleration bias. Therefore, the process noise can be defined as below.
\begin{equation} \label{equ_acc_Q}
\begin{split}
    Q_{k-1} &= 
    \begin{bmatrix}
        Q'_{k-1} & 0 & 0 \\
        0 & Q'_{k-1} & 0 \\
        0 & 0 & Q'_{k-1}
    \end{bmatrix} \\
    Q'_{k-1} &= 
    \begin{bmatrix}
        \frac{T^3 \tau_a}{3}+\frac{T^5 \tau_b}{20} & \frac{T^2 \tau_a}{2}+\frac{T^4 \tau_b}{8} & -\frac{T^3 \tau_b}{6} \\
        \frac{T^2 \tau_a}{2}+\frac{T^4 \tau_b}{8}  & T\tau_a+\frac{T^3 \tau_b}{3}              & -\frac{T^2 \tau_b}{2} \\
        -\frac{T^3 \tau_b}{6}                      & -\frac{T^2 \tau_b}{2}                     & T \tau_b
    \end{bmatrix}
\end{split}
\end{equation}
The measurement update is almost identical with (\ref{equ_measurement_update}), except that the linearized measurement matrix is replaced by (\ref{equ_acc_H}).
\begin{equation} \label{equ_acc_H}
H_k =
\begin{bmatrix}
\frac{p_x-p_{Ax}}{\bar{r}_k} & 0 & 0 & \frac{p_y-p_{Ay}}{\bar{r}_k} & 0 & 0 & \frac{p_z-p_{Az}}{\bar{r}_k} & 0 & 0
\end{bmatrix}^T
\end{equation}

\subsection{Implementation Details}
Until now, the above EKF with augmented state vector is completed. The acceleration from IMU is fused into UWB measurements as the control input, while the acceleration bias is estimated as part of the state vector. But in practice, the readings from UWB sensors are unstable. In most situations, the measured distance $r_k$ is accurate up to a few centimeters, while occasionally the reported distance is completely wrong, e.g. a few meters away from the ground truth. In addition, the IMU readings are quite noisy, which means that relying too much on the IMU is impractical as well. Usually we tend to rely more on UWB readings when tuning the covariance matrices $Q$ and $R$, with the price that the jumping of UWB readings will result in sudden change of the estimated position.

To alleviate such problems, we compute the difference between the predicted distance $\bar{r}_k$ and the actual UWB measurements $r_k$, $d_k = |\bar{r}_k - r_k|$. If $d_k$ is over a certain threshold, e.g. 2m, the localization result $x_k$ is discarded.
\section{Experiments} \label{sec_experiments}
\begin{figure}[t!] \centering
    \includegraphics[width=0.9\columnwidth]{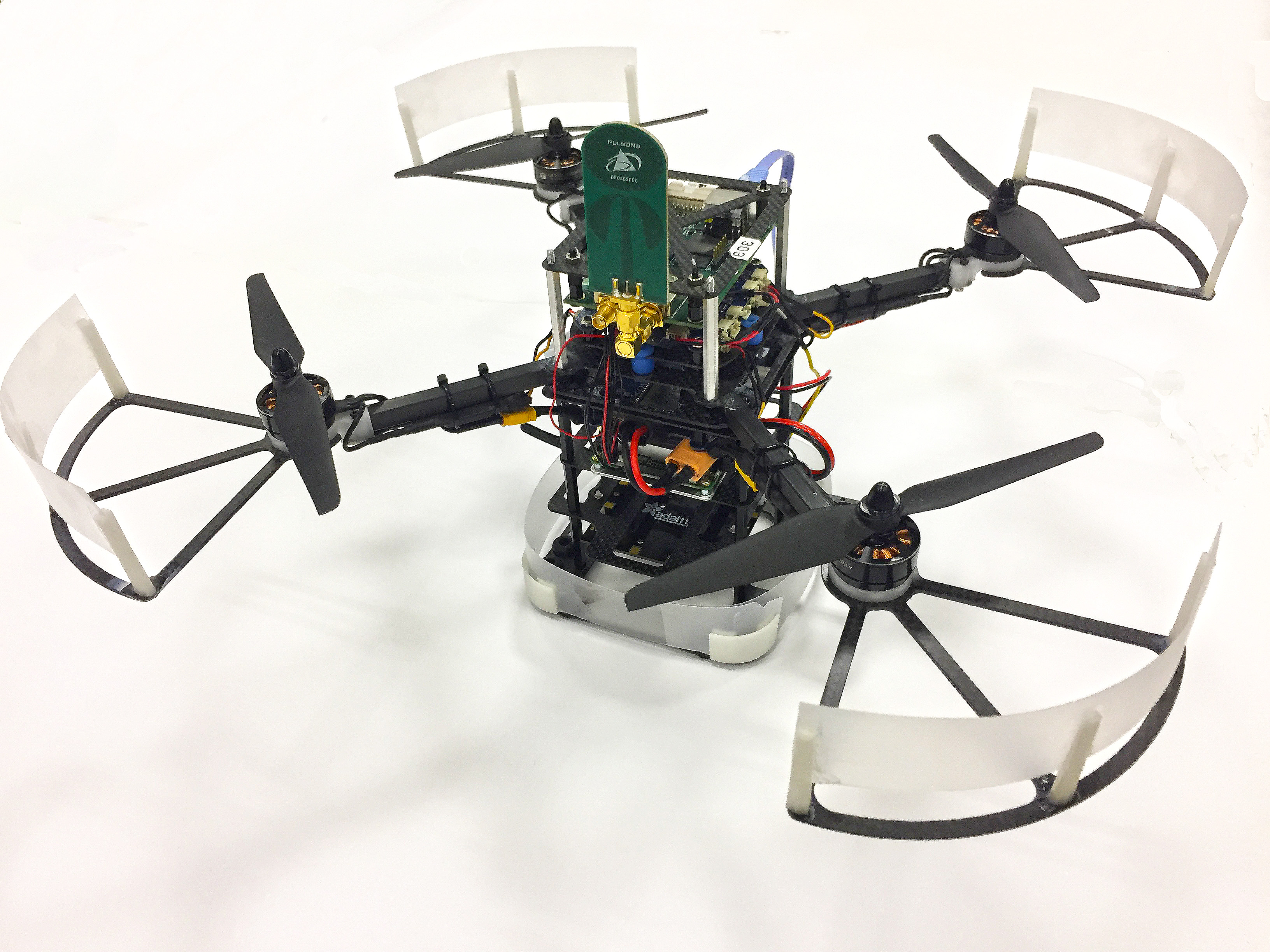}
    \caption{The quadrotor platform}\label{fig_quad}
\end{figure}
\begin{figure*}[t!] 
\centering
  \parbox{\figrasterwd}{
    \parbox{.49\figrasterwd}{
      \subcaptionbox{}{\includegraphics[width=\hsize]{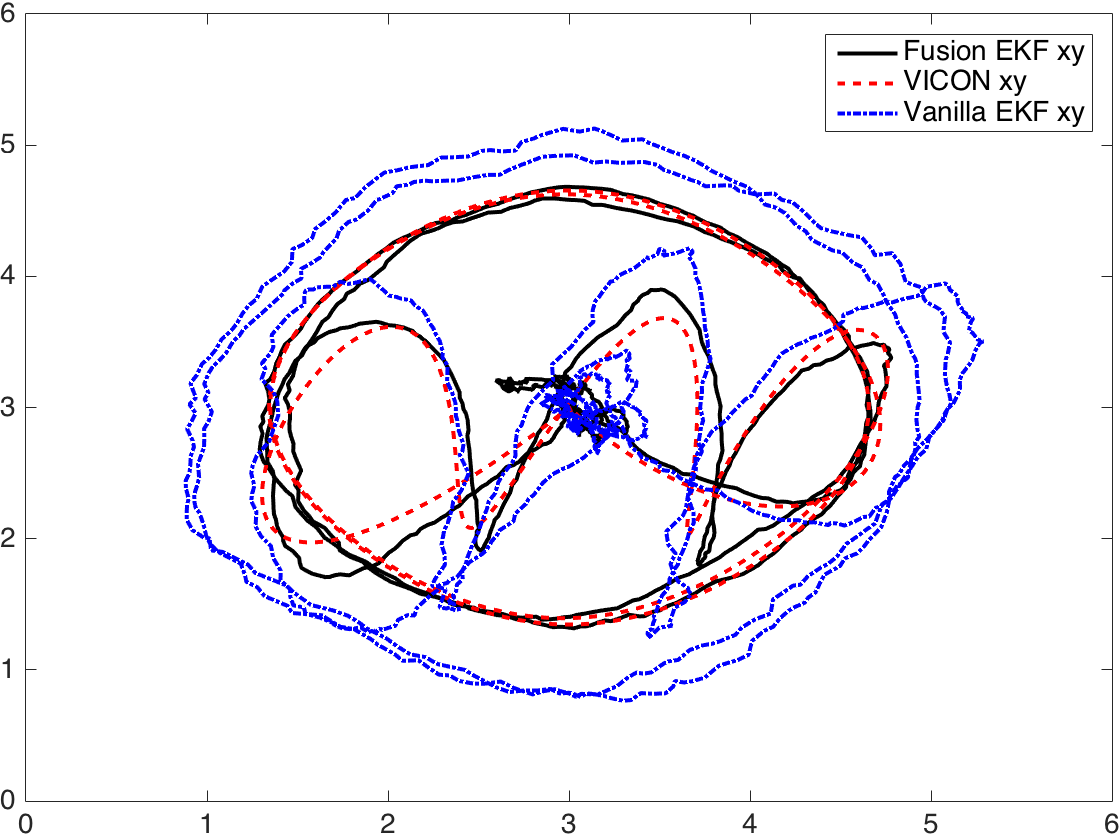}} \label{fig_vicon_xy}
    }
    \parbox{.24\figrasterwd}{
      \subcaptionbox{}{\includegraphics[width=\hsize]{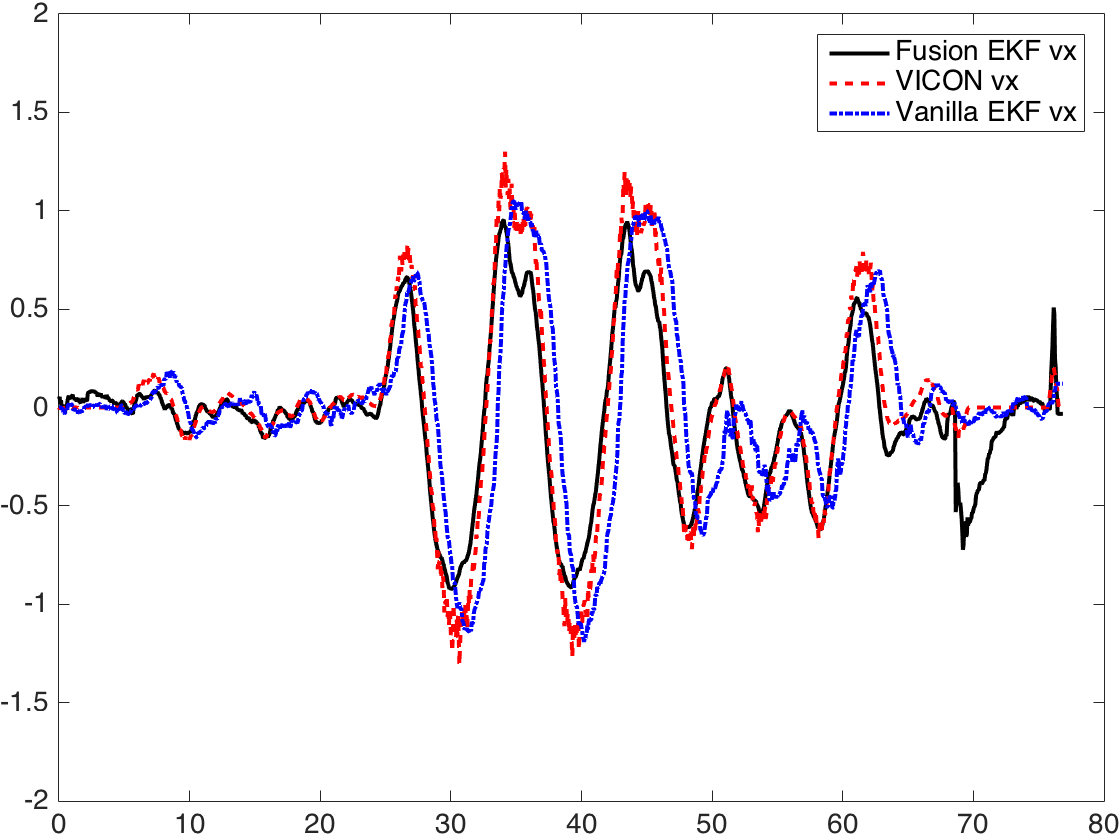}}\label{fig_vicon_vx}
      \subcaptionbox{}{\includegraphics[width=\hsize]{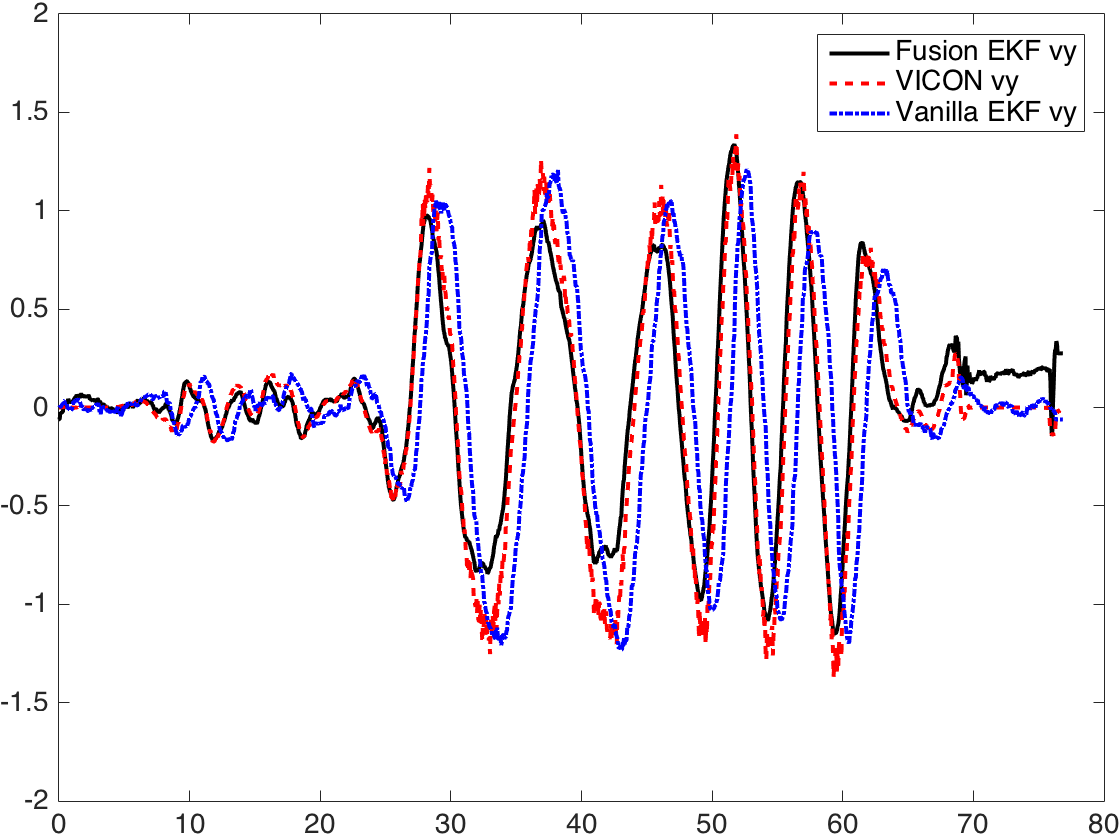}} \label{fig_vicon_vy}
    }
    \parbox{.24\figrasterwd}{
      \subcaptionbox{}{ \includegraphics[width=\hsize]{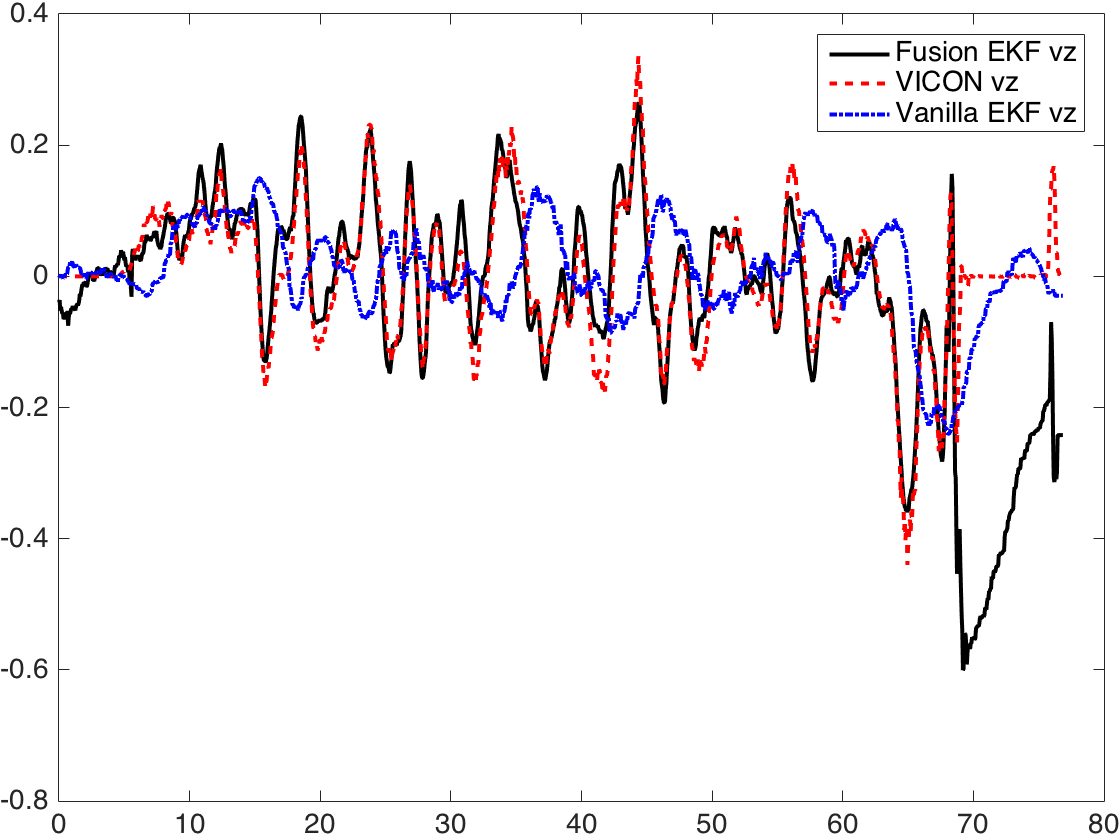}}\label{fig_vicon_vz}
      \subcaptionbox{}{\includegraphics[width=\hsize]{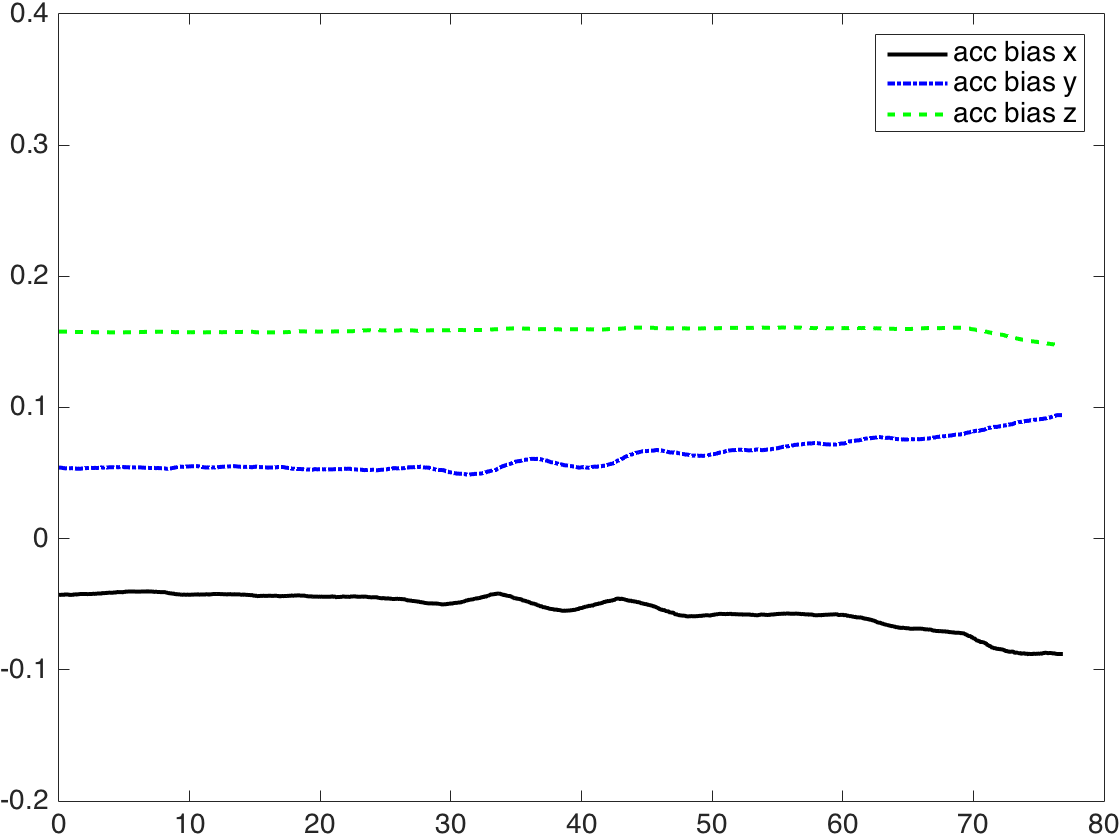}}\label{fig_vicon_bias}
    }
  }
  \caption{(a) The trajectory in x-y plane. (b-d) The MAV velocity along the axis of x,y,z. (e) The estimated acceleration bias of the IMU. }\label{fig_vicon_1}
  
\end{figure*}
\begin{figure*}[h!] 
    \parbox{.49\figrasterwd}{
      \subcaptionbox{}{\includegraphics[width=\hsize]{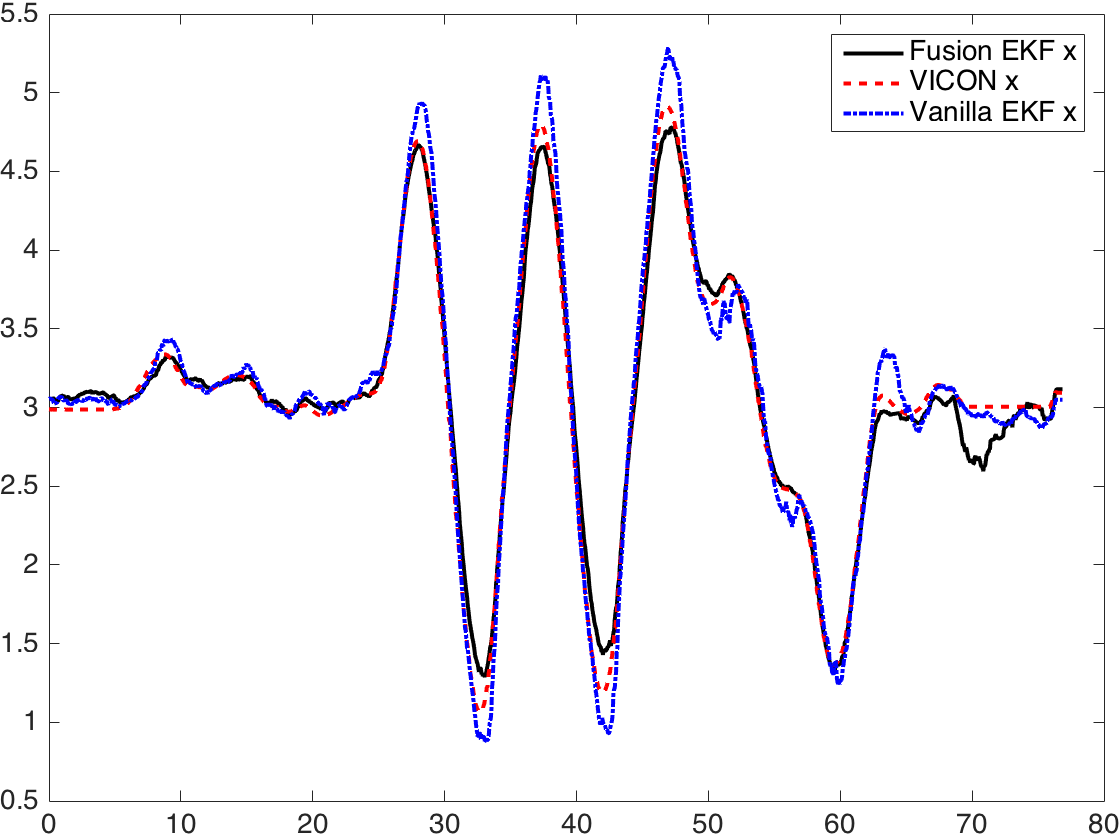}}\label{fig_vicon_x}
      \subcaptionbox{}{\includegraphics[width=\hsize]{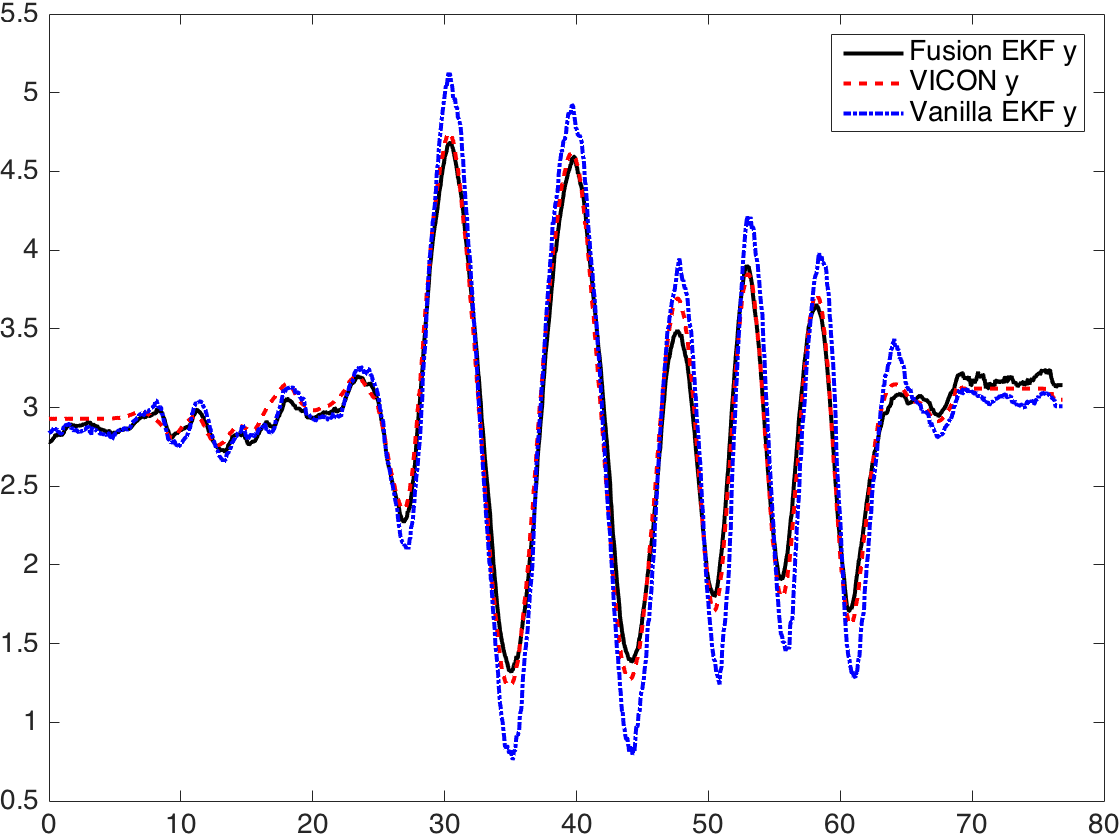}} \label{fig_vicon_y}
    }
    \parbox{.49\figrasterwd}{
      \subcaptionbox{}{\includegraphics[width=\hsize]{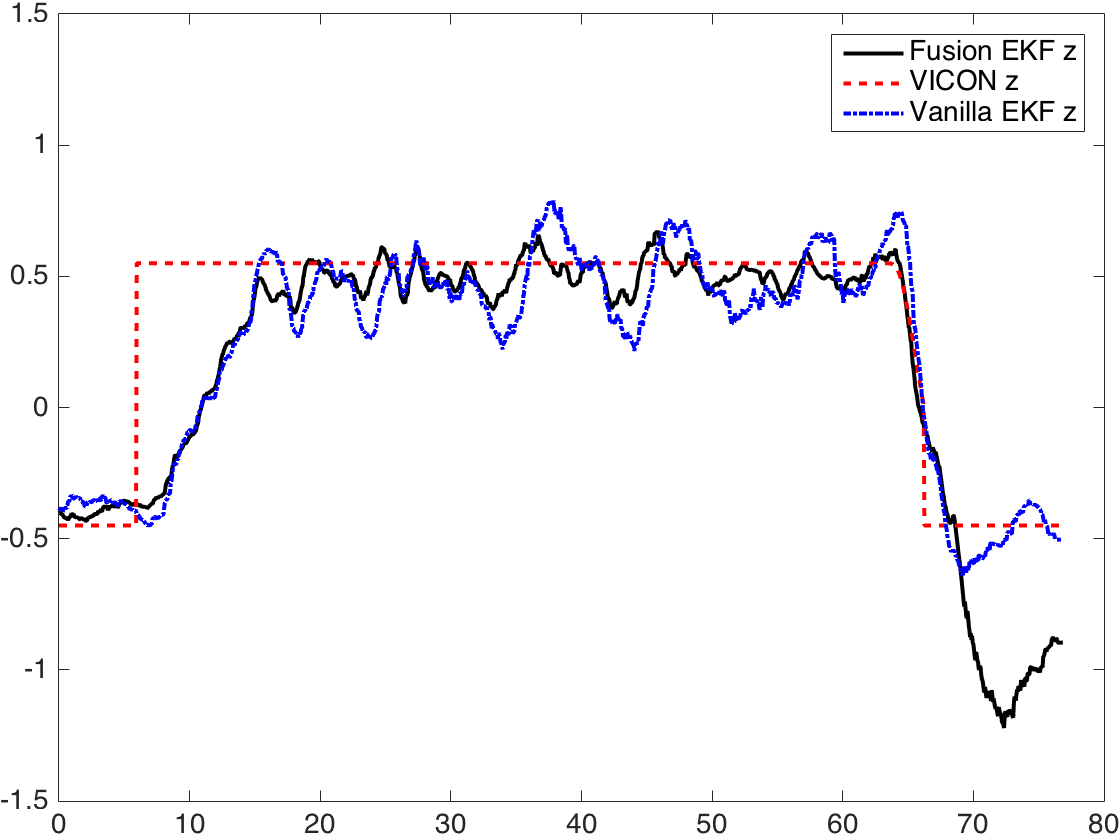}}\label{fig_vicon_z}
      \subcaptionbox{}{\includegraphics[width=\hsize]{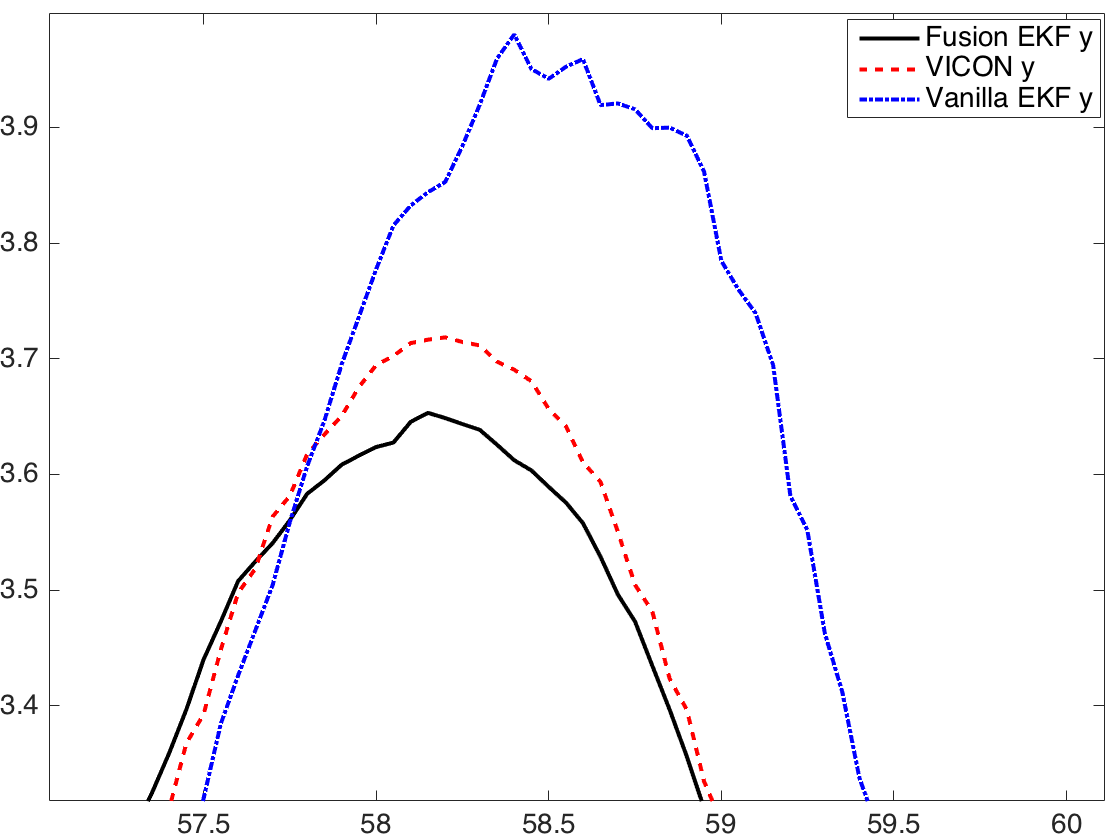}}  \label{fig_vicon_y_enlarged}
    }
    \caption{(a-c) the MAV position along the axis of x,y,z. (d) A closer look into (b). It can be observed that the delay between VICON measurement and fusion EKF estimation is nearly zero, while the delay between vanilla EKF estimation and ground truth is about 0.5s.}\label{fig_vicon_2}
\end{figure*}
First of all, the vanilla EKF in Section \ref{sec_vanilla_EKF} and the fusion EKF in Section \ref{sec_fusion_EKF} are tested in an environment equipped with VICON system, so that ground truth is available and the performance of the two algorithms can be quantitatively evaluated. Shown in Figure \ref{fig_vicon_1} and \ref{fig_vicon_2}, it is obvious that the fusion EKF exhibits significantly better accuracy and much lower latency. The fusion EKF is deployed on 6 MAVs that performs a swarming light show at Changi Exhibition Centre, Singapore. This light show is the key performance at the ceremony of two exhibitions, namely Unmanned System Asia 2017 and Rotorcraft Asia 2017. 

The MAV platform is shown in Fig. \ref{fig_quad}. The flight controller is an enhanced version of Pixhawk. The fight controller, sensors like magnetometer, IMU, barometer, regulated power supplier are integrated into a single compact circuit. The UWB sensors are commercial products provided by the company of TimeDomain. As mentioned in Section \ref{sec_sensor_setup}, the UWB sensors are configured to work at two-way ranging mode with TDMA and CDMA. They provide distance measurements between mobiles and anchors at the speed of about 80Hz, and the typical accuracy is 10cm. The acceleration measurements come from the onboard IMU sensor at the frequency of 50Hz. An upboard, i.e. a commercial Intel Atom based microcomputer, is mounted on the MAV to host the mission management and localization algorithm. The total weight of the MAV is 800g, including a 2200mAh 3-cell battery, a high power LED array, and blade protectors.

In both test at VICON room and the exhibition hall, the 6 MAVs execute a pre-defined path designed by a multi-agent splines based trajectory generation algorithm \cite{lai2016robust}. The maximum velocity of the MAVs is $1.2m/s$, and the maximum acceleration is $2m/s^2$. In experiments presented in this paper, only one of the MAVs are used, while the other 5 MAVs differs only on the path executed. The positioning of MAVs in Section \ref{sec_experiments} relies on the fusion EKF algorithm.

\subsection{VICON Test}
In the room equipped with VICON system, the UWB anchors are placed on the ground and the ceiling. Limited by the size of VICON room, the maximum distance between anchors is about 6m at horizontal direction, and 2.5m at vertical direction. Actually such setup with nearby anchors significantly downgrades the performance of the UWB localization system. The error of measured distance is around 10cm given that UWB sensors are within maximum working distance, which is a few hundred meters. The error is at the same level no matter the distance between sensors as long as the distance is within the maximum working distance. Therefore, in a compact setup, the ratio between measurement error and measured distance is larger, which makes triangulation less accurate. Given such a small setup at VICON room, the performance is not ideal but still enough for the MAV to execute a path. 

The results of executing a pre-defined path is shown in Figure \ref{fig_vicon_1} and \ref{fig_vicon_2}. According to the position and velocity curves along x,y,z direction (NWU), the fusion EKF have significantly higher accuracy than that of vanilla EKF. In particular, the vanilla EKF suffers from overshooting, which is as expected because of the constant velocity assumption. Moreover, compared to the accuracy problem, the delay exhibited in vanilla EKF is fatal. The latency between estimated position and VICON measurement is about 500ms in Figure \ref{fig_vicon_2}, which is large enough to diverge the controller of the MAV. The maximum and average positioning errors are demonstrated in Table \ref{tbl_result}.

\begin{table}[t!]
\caption{Performance of Fusion EKF and Vanilla EKF}
\centering
\label{tbl_result}
\addtolength{\tabcolsep}{-1pt}  
\begin{tabular}{*{3}c}
 \hline
   & Maximum Error (m) & Mean Error (m) \\
 \hline
 Fusion EKF & 0.39 & 0.16 \\
 Vanilla EKF & 0.71 & 0.30 \\
 \hline
\end{tabular}
\end{table}

Although it is difficult to get the ground truth of the IMU acceleration bias, the scale and trend of the estimated bias in Figure \ref{fig_vicon_1} seems reasonable. Based on the accuracy of the localization result, we can safely assume that the estimated bias is roughly correct. Another phenomenon is that, for the fusion EKF, the estimated position and velocity dropped significantly at about 68s. That's because the MAV lands on the ground at that time, and the hit introduces a huge acceleration measurement from the IMU. Since the system matrix of fusion EKF relies on the acceleration measurement, such abnormal input will lead to incorrect estimation. But generally such phenomenon will not cause any problem as it happens only at landing or MAV's crashing into something.

\subsection{Performance at Changi Exhibition Centre}
\begin{figure*}[t!] 
\centering
  \parbox{\figrasterwd}{
    \parbox{.49\figrasterwd}{
      \subcaptionbox{}{\includegraphics[width=\hsize]{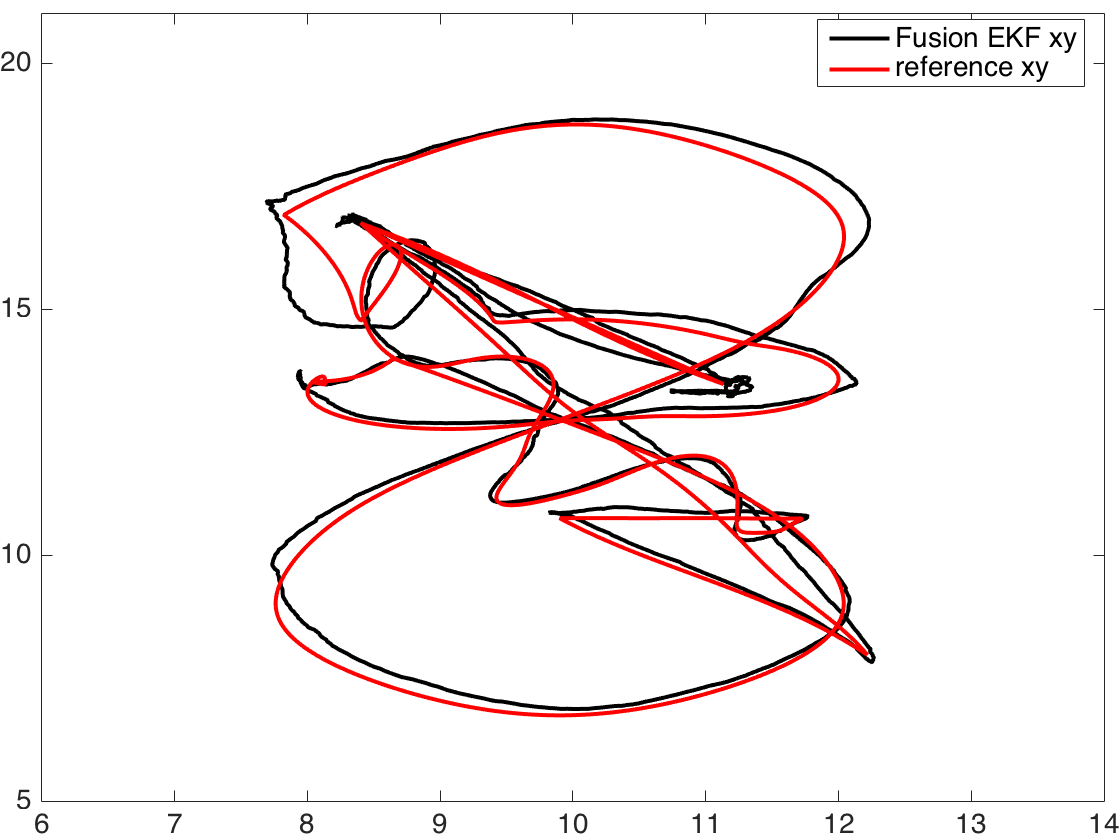}} 
    }
    \parbox{.24\figrasterwd}{
      \subcaptionbox{}{\includegraphics[width=\hsize]{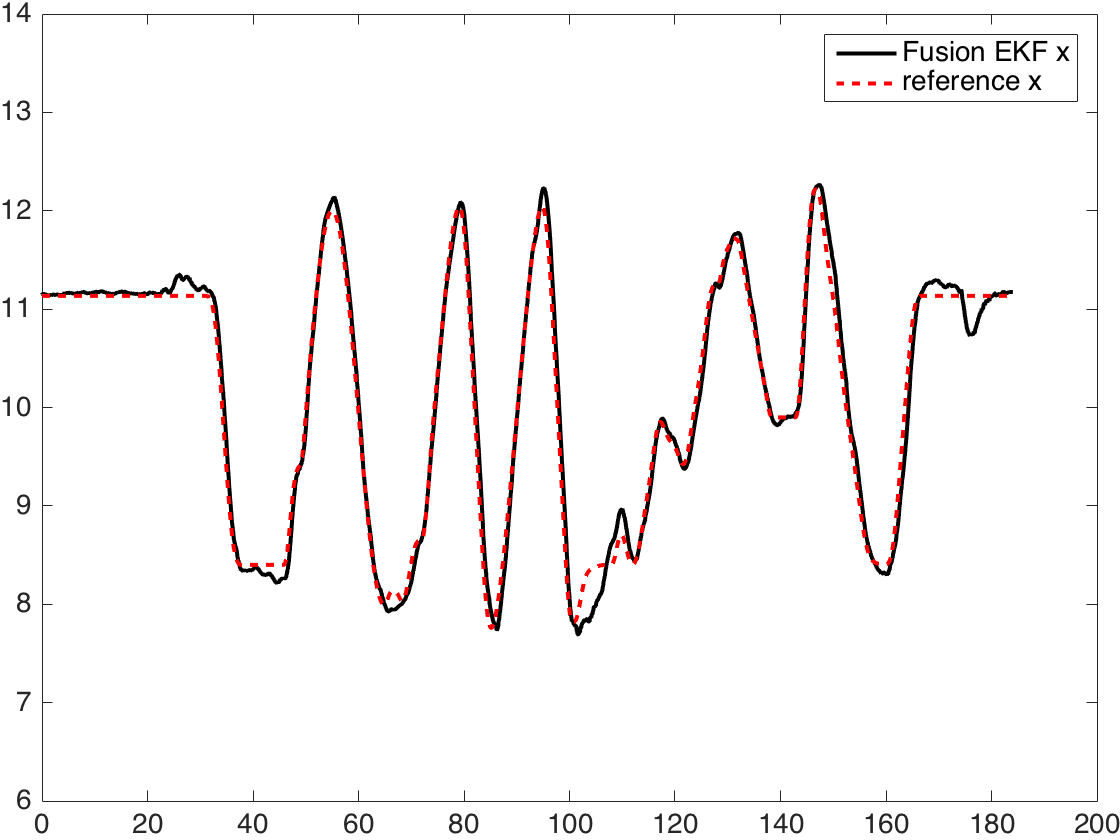}}
      \subcaptionbox{}{\includegraphics[width=\hsize]{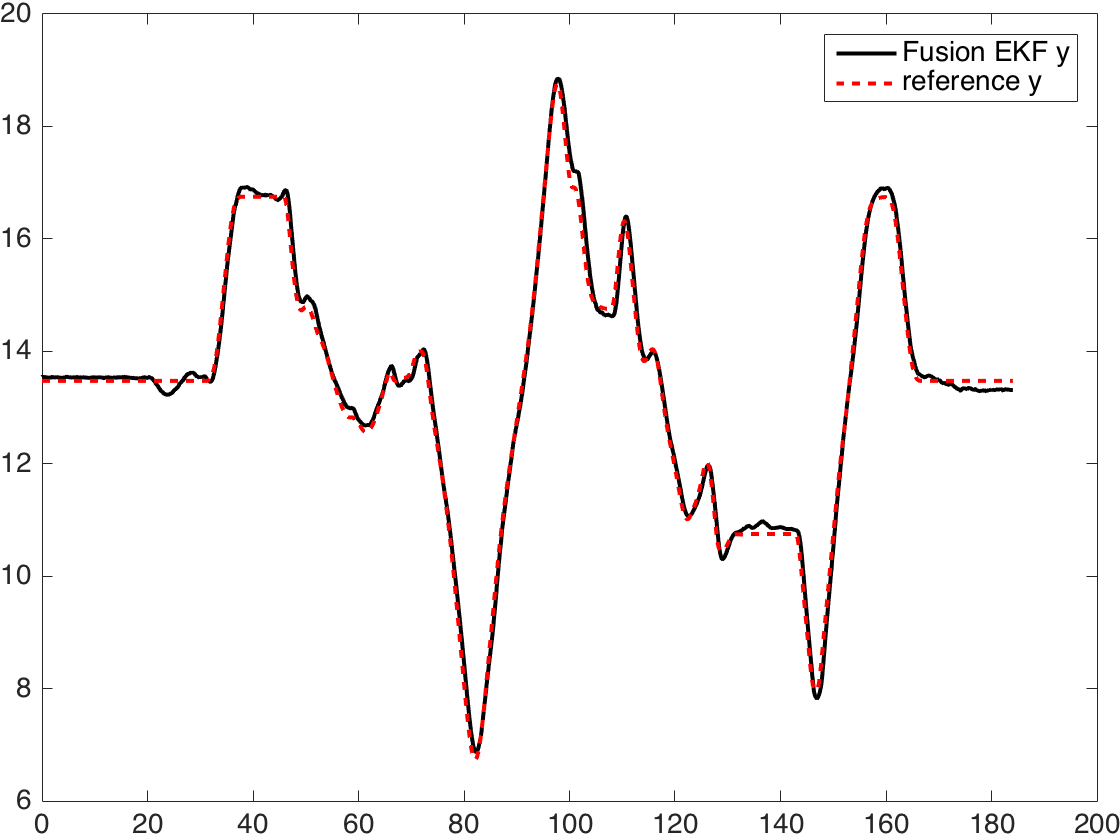}}
    }
    \parbox{.24\figrasterwd}{
      \subcaptionbox{}{ \includegraphics[width=\hsize]{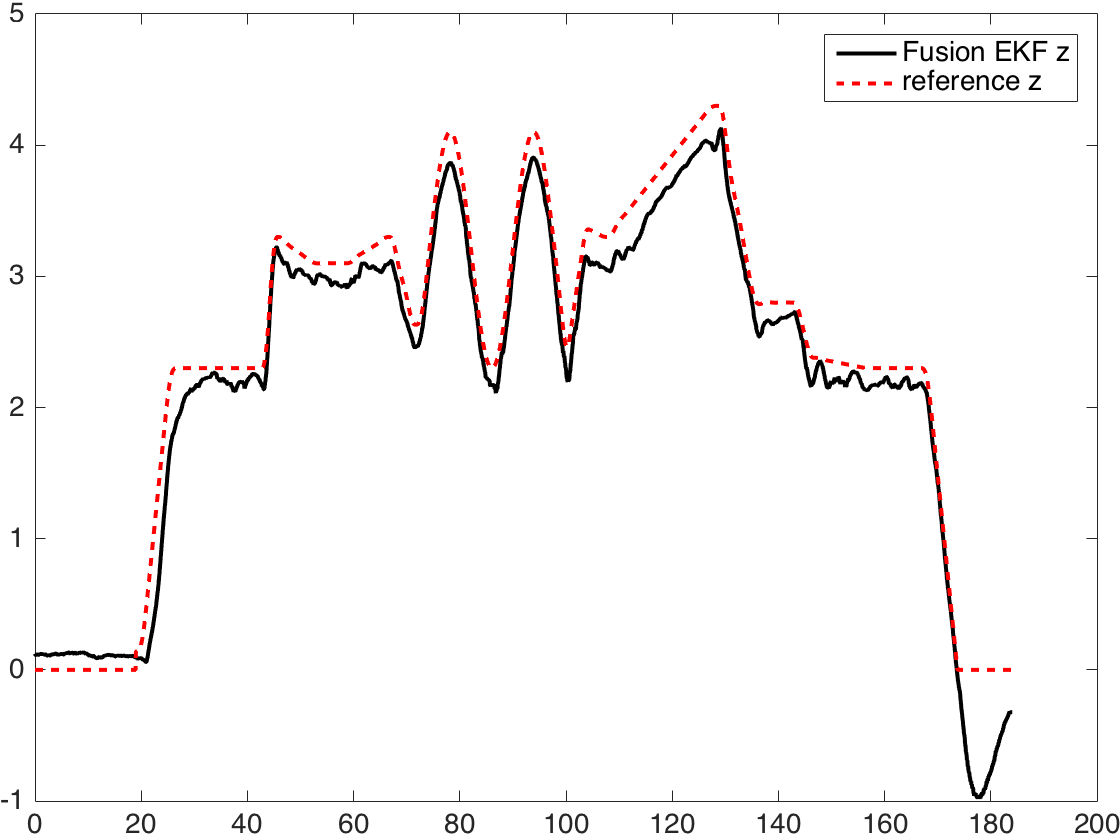}}
      \subcaptionbox{}{\includegraphics[width=\hsize]{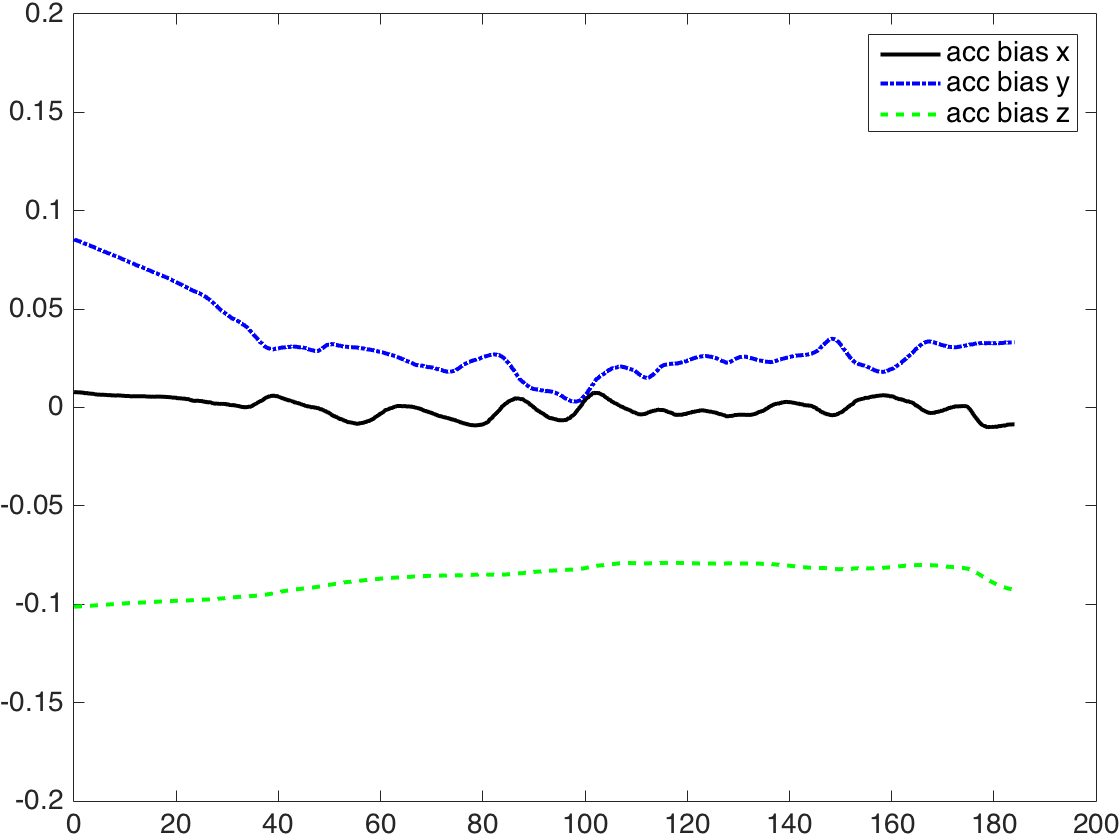}}
    }
  }
  \caption{(a) The trajectory in x-y plane. (b-d) The MAV position along the axis of x,y,z. (e) The estimated acceleration bias of the IMU. }\label{fig_418}
  
\end{figure*}
Utilizing the fusion EKF, a MAV swarm performance is conducted in Changi Exhibition Centre, Singapore, on the ceremony of Unmanned System Asia 2017 and Rotorcraft Asia 2017. In the performance, the UWB anchors are setup in a much larger space shown in Figure \ref{fig_anchor}. As mentioned above, a setup with larger space between UWB sensors leads to more stable and accurate estimation. Actually the MAVs are able to hover without visible movement in the exhibition hall. 

But in the case of multiple MAV operation, other difficulties emerge. First of all, the electromagnetic was much more complex with 12 UWB sensors working in full power, as well as thousands of WiFi devices around. Secondly, in a MAV formation, the MAVs block the UWB signals of others. The nature of UWB ranging requires complete line-of-sight environment between any pair of sensors. Any kind of occlusion may leads to totally incorrect readings. Both reasons resulted in larger ranging error, and more frequent measurement jumping. In our proposed fusion EKF, apparently incorrect UWB readings are rejected by the innovation thresholding. In the case of thresholding failure, the integration with acceleration keeps the estimated position from deviating too much from ground truth.

Shown in Figure \ref{fig_418}, the localization result is clean and close to the reference. Although there is no way to get the ground truth in such a public performance, the localization accuracy can be evaluated via videos at \url{https://youtu.be/1id49danIK4}.

\section{Conclusion}
In this paper, an EKF based algorithm is proposed to fuse the measurements of UWB sensors and IMU. The position, velocity, and the bias of the IMU are estimated simultaneously. Experiments with VICON system have proved that the stability and accuracy of our approach significantly outperform vanilla EKF. More importantly, the delay of the localization algorithm is almost eliminated while vanilla EKF exhibits unacceptable estimation delay. Further more, a multi-MAV light show is performed at an indoor exhibition hall, using the positioning algorithm proposed in this paper.
%


\footnotesize
\bibliographystyle{IEEEtran}
\bibliography{ICCA2018_REF}




\end{document}